\documentclass[final]{cvpr}

\usepackage{times}
\usepackage{epsfig}
\usepackage{graphicx}
\usepackage{amsmath}
\usepackage{amssymb}

\usepackage{xspace}
\usepackage{amsmath,amsfonts,amssymb,amsthm}
\usepackage{bm}
\usepackage[dvipsnames]{xcolor}
\usepackage{diagbox}
\usepackage{pifont}
\usepackage{amsmath}
\usepackage{enumitem}
\usepackage{mathtools}
\usepackage{algorithm}
\usepackage{algorithmic}
\usepackage{gensymb}
\usepackage{wrapfig}
\usepackage{subcaption}
\usepackage{multirow}
\usepackage{bbm}
\usepackage{colortbl}
\usepackage{textcomp}

\usepackage[pagebackref=true,breaklinks=true,colorlinks,bookmarks=false]{hyperref}

\def\cvprPaperID{****} % *** Enter the CVPR Paper ID here
\def\confYear{CVPR 2021}
\newcommand{\cmark}{\ding{51}}%

\DeclarePairedDelimiter{\norm}{\lVert}{\rVert}

\DeclareMathOperator*{\argmin}{arg\,min}

\newcommand{\indep}{\perp \!\!\! \perp}

\begin{document}

%%%%%%%%% TITLE
\title{Temporal Action Detection with Multi-level Supervision}

\author{Baifeng Shi\\
Peking University
% {\tt\small bfshi@pku.edu.cn}
\and
Qi Dai\\
Microsoft Research Asia
% {\tt\small qid@microsoft.com}
\and
Judy Hoffman\\
Georgia Tech
% {\tt\small judy@gatech.edu}
\and
Kate Saenko\\
Boston University \& MIT-IBM Watson AI Lab
% {\tt\small saenko@bu.edu}
\and
Trevor Darrell, 
Huijuan Xu\\
UC Berkeley
% {\tt\small \{trevor,huijuan\}@eecs.berkeley.edu}
}

\maketitle

%%%%%%%%% ABSTRACT
\begin{abstract}
Training temporal action detection in videos requires large amounts of labeled data, yet such annotation is expensive to collect. Incorporating unlabeled or weakly-labeled data to train action detection model could help reduce annotation cost. In this work, we first introduce the Semi-supervised Action Detection (SSAD) task with a mixture of labeled and unlabeled data and analyze different types of errors in the proposed SSAD baselines which are directly adapted from the semi-supervised classification task. To alleviate the main error of action incompleteness (\ie, missing parts of actions) in SSAD baselines, we further design an unsupervised foreground attention (UFA) module utilizing the ``independence'' between foreground and background motion. Then we incorporate weakly-labeled data into SSAD and propose Omni-supervised Action Detection (OSAD) with three levels of supervision. An information bottleneck (IB) suppressing the scene information in non-action frames while preserving the action information is designed to help overcome the accompanying action-context confusion problem in OSAD baselines. We extensively benchmark against the baselines for SSAD and OSAD on our created data splits in THUMOS14 and ActivityNet1.2, and demonstrate the effectiveness of the proposed UFA and IB methods. Lastly, the benefit of our full OSAD-IB model under limited annotation budgets is shown by exploring the optimal annotation strategy for labeled, unlabeled and weakly-labeled data.
\end{abstract}

\section{Introduction}

%Action detection is one of the most fundamental tasks in video understanding~\cite{zhao2017temporal,xu2017r,lin2018bsn,shou2017cdc,chao2018rethinking} where one needs to predict both the category and temporal location of the actions in a video. Recent success of action detection models highly depends on large amount of fully-labeled training data with both classification and localization annotations. However, the annotation process, especially for localization, is extremely time-consuming. This requires us to develop algorithms to improve current action detection models with unlabeled data which barely needs any cost to collect. Although learning from unlabeled data has long been studied in the task of image classification, it is still unexplored in the action detection problem. In this work, we fill the gaps by learning action detection with unlabeled videos together with other levels of supervision.
Temporal action detection is one of the most fundamental tasks in video understanding, which requires simultaneously classifying the actions in a video and localizing their start and end times. Recent success of temporal action detection models~\cite{zhao2017temporal,xu2017r,lin2018bsn,shou2017cdc,chao2018rethinking} highly relies on large amounts of fully-labeled training data with both classification and localization annotations. However, the annotation process, especially for localization, is extremely time-consuming and expensive. 
To alleviate this problem, one direction is to maximize the usage of unlabeled or weakly-labeled data to bring performance improvement with less annotation cost. In this work, we study temporal action detection using fewer labeled videos together with other levels of supervision, e.g. unlabeled and weakly-labeled videos.

% Previous works have explored the task of weakly-supervised action detection (WSAD)~\cite{nguyen2018weakly,nguyen2019weakly,shi2020weakly,paul2018w,liu2019completeness,luo2020weakly} where only the video-level category is available. 

\begin{figure}[t]
\centering
\includegraphics[width=\linewidth]{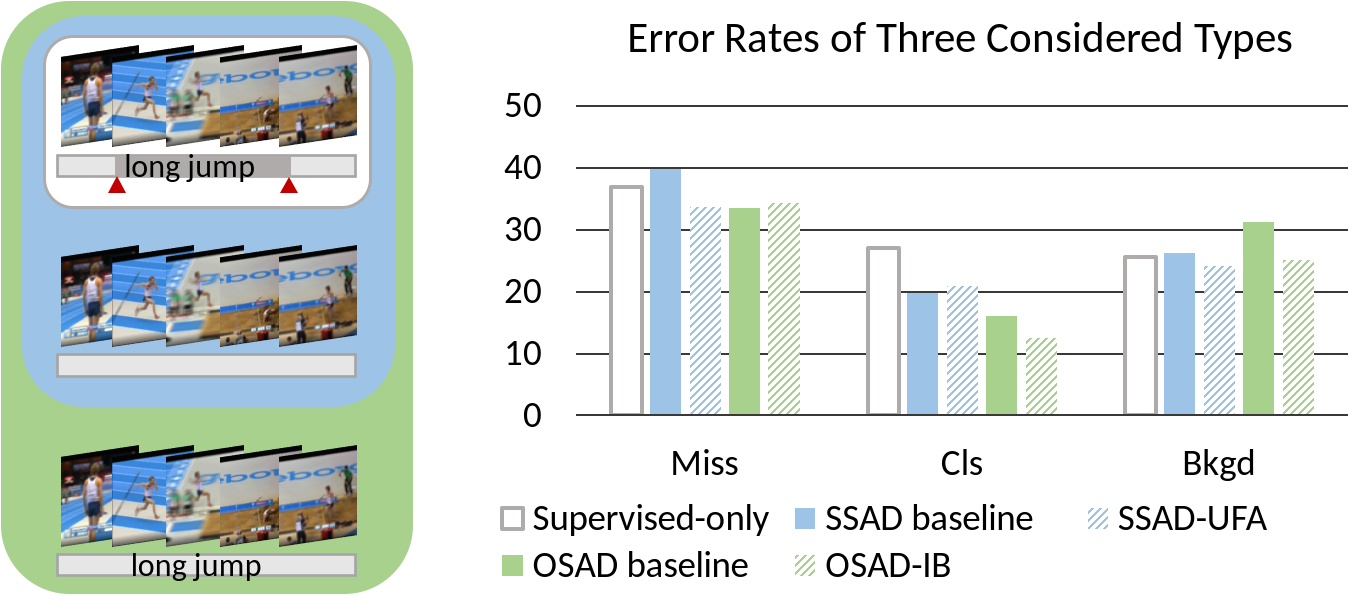}
\caption{\small \emph{Left}: Data type for different tasks. FSAD uses only fully-labeled videos (white area). SSAD uses both fully-labeled and unlabeled videos (blue area). OSAD further uses weakly-labeled videos besides these two data types (green area). \emph{Right}: Error analysis on SSAD/OSAD models. We consider three types of errors: 1) Action incompleteness (Miss), 2) Misclassification (Cls), and 3) Action-context confusion (Bkgd). 
   Miss is the main type of error in SSAD baseline, while Bkgd is the main type of error in OSAD baseline.
   See Sec.~\ref{sec:implementation} for more details.}
\label{fig:intro}
\end{figure}

Learning from unlabeled data has been investigated in the task of semi-supervised image classification~\cite{oliver2018realistic,tarvainen2017mean,berthelot2019mixmatch,sohn2020fixmatch} and shows promising results, while such problem setting is unexplored in the temporal action detection.
We introduce the Semi-supervised Action Detection (SSAD) task and establish three SSAD baselines by incorporating three state-of-the-art Semi-Supervised Learning (SSL) models (Mean Teacher~\cite{tarvainen2017mean}, MixMatch~\cite{berthelot2019mixmatch}, FixMatch~\cite{sohn2020fixmatch}) into a Fully-Supervised Action Detection (FSAD) backbone. 
For the purpose of \textbf{initial} evaluation on SSAD benchmark, we choose a straightforward yet effective FSAD method, SSN~\cite{zhao2017temporal}, as our backbone and leave the development of more complex backbones for future work.
%Since this is the \textbf{initial} evaluation on SSAD benchmark, we choose a straightforward yet effective FSAD method, SSN~\cite{zhao2017temporal}, as our baseline and leave the development of more complex methods to future work.
% \huijuan{maybe, we don't need to talk about SSN here, just mention it in the experiment section.}
However, directly applying SSL algorithms in the SSAD baselines only brings small result improvement compared to the supervised-only model. 
To track down the main source of error, we conduct error analysis for the SSAD baselines (Fig.~\ref{fig:intro})
and find the main problem of the SSAD baseline is \textbf{action incompleteness}, namely missing parts of the action duration.
%\huijuan{Here in the introduction, we simply refer to figure 1 and give the main take-away...So you may refer to figure 1 again in the experiment section in more detail...}
%Here we consider 3 common Types of errors: 1) \textbf{Action incompleteness}: missing parts of action; 2) \textbf{Misclassification}: incorrect action classification; 3) \textbf{Action-context confusion}: recognizing non-action frames as action. As shown in Table~\ref{table:error_analysis}, SSAD baselines have lower error rate of Type-3, while having more errors of Type-1, implying they have trouble recognizing the action when it happens. 
To prevent the SSAD baseline from ignoring action frames, we borrow the idea from object-centric representations~\cite{fathi2013modeling,mccandless2013object,wang2018videos,materzynska2020something} to extract more discriminative representation of action which is basically characterized by the foreground objects (humans).\footnote{In this paper we use the terms ``background'' and ``foreground'' to indicate the spatial regions in each frame, and ``action'' and ``non-action'' denote the temporal frames.} There have been attempts to endow machines with the ability to detect salient moving objects. However, they either need manual annotation~\cite{tokmakov2017learning,tokmakov2017learningb,song2018pyramid,cheng2017segflow}, or make assumptions improper for action videos~\cite{yang2019unsupervised}. In this work, we propose to detect the foreground without supervision by leveraging the ``independence" between foreground and background motions, \ie, the foreground motion is self-contained and not affected by the motion of background. 
Specifically, we learn the attention by minimizing the information reduction rate between foreground and background motion in unlabeled data.
To this end, our proposed unsupervised foreground attention (UFA) module successfully helps SSAD models recognize relatively complete actions without extra annotation cost.

Further, we consider weakly-labeled data with only video-level category labels which is in the middle of the cost-accuracy trade-off between fully-labeled and unlabeled data.
%In another line of work, the task of Weakly-Supervised Action Detection (WSAD) is also explored~\cite{nguyen2018weakly,nguyen2019weakly,shi2020weakly,paul2018w,liu2019completeness,luo2020weakly}, where only the video-level category is labeled. 
It has been shown that weakly-supervised temporal action detection~\cite{nguyen2018weakly,nguyen2019weakly,shi2020weakly,paul2018w,liu2019completeness,luo2020weakly} can save annotation cost without degrading the performance too much. 
Therefore, we further include weakly-labeled data into the SSAD model and form a unified framework with three levels of supervision, named Omni-supervised Action Detection (OSAD). 
As a baseline for OSAD, we simply add a video-level classification loss for the additional weakly-labeled data. However, training video-level classification to realize weak action localization %\huijuan{say more.... e.g. to realize temporal localization based on class activation map....} 
may cause \textbf{action-context confusion}~\cite{liu2019completeness,shi2020weakly}, \ie, the model is highly activated at non-action frames because they contain background \emph{scene information} (\eg swimming pool) which is highly indicative of the action category (\eg swimming). This phenomenon is also verified 
in our error analysis (Fig.~\ref{fig:intro}) where the OSAD baseline has higher action-context confusion and incorrectly recognizes non-action frames as action.
To alleviate the issue, we propose an information bottleneck (IB) method to filter out the scene information extracted from non-action frames while preserving the action information by training action classification. Specifically, we regularize the entropy of the features of non-action frames which only contain scene information, thus significantly reduce action-context confusion in our full OSAD-IB model (Fig.~\ref{fig:intro}).

We conduct extensive experiments on SSAD and OSAD baselines, as well as the proposed UFA and IB. Moreover, we show the advantage of multi-level supervision over single-level supervision in a realistic scenario, where we search the best labeling policy under an annotation budget. 

To sum up our main contributions, we: (i) propose the SSAD and OSAD tasks to utilize unlabeled and weakly-labeled data in temporal action detection, and establish several baseline models for them; (ii) design an unsupervised foreground attention module to alleviate the action incompleteness problem in SSAD baselines; (iii) design an information bottleneck method to solve the action-context confusion problem in OSAD baselines; (iv) validate the proposed SSAD and OSAD methods through extensive experiments, and show the advantage of our full OSAD-IB model under a realistic scenario where an annotation budget is given.

%To sum up, our main contributions include: (i) The first baseline models for SSAD and OSAD tasks; (ii) An unsupervised foreground attention module to alleviate the action incompleteness problem in SSAD baselines; (iii) An information bottleneck method to solve the action-context problem in OSAD baselines; (iv) Extensive experiments on SSAD and OSAD, under a realistic scenario where a fixed annotation budget is given.
\section{Related Work}

\textbf{Fully-supervised Action Detection}.
The basic paradigms in FSAD methods share significant similarities with their counterparts in the object detection area~\cite{girshick2014rich,girshick2015fast,ren2015faster,redmon2016you,liu2016ssd}. The most common one is the two-stage pipeline, where proposals are first uniformly sampled from a video and then classified and re-localized. %The model is trained with a classification loss and a regression loss. 
R-C3D~\cite{xu2017r} improves the proposal quality with a proposal subnet. SSN~\cite{zhao2017temporal} optimizes a completeness loss to avoid the incomplete proposals. It also proposes the feature pyramid (STPP) to better capture the temporal structure, and the temporal actionness grouping (TAG) for proposal sampling. TAL-Net~\cite{chao2018rethinking} borrows the architecture of Faster R-CNN~\cite{ren2015faster} in object detection and make it accommodate the action detection setting by multiple modifications. Recent works also focus on refining the temporal structures, \eg, with multiple temporal scales~\cite{shou2016temporal}, higher temporal resolution~\cite{shou2017cdc}, precise temporal boundaries~\cite{lin2018bsn}, or using graph networks to model the temporal relations~\cite{xu2020g,zeng2019graph}. In this work, we choose SSN as the FSAD base model. Note that we discard STPP and TAG, and only keep the completeness loss to simplify the algorithm without hurting the performance too much.

\textbf{Semi-supervised Learning}.
%A main line of work in SSL can be coarsely 
Most semi-supervised learning methods can be summarized as training the model to predict the pseudo label or produce a consistent output of the input with different augmentations. Pseudo label (or self-training) methods~\cite{lee2013pseudo} require to convert model outputs into hard pseudo labels using a sharpening function and encourage the model to predict the pseudo labels with high confidence. Consistency-based methods~\cite{laine2016temporal,tarvainen2017mean,oliver2018realistic} exploit the output of themselves or their time-average version as ``soft label,'' and make the model generate a consistent output when the input is randomly or adversarially augmented. FixMatch~\cite{sohn2020fixmatch} combines pseudo-label and consistency-based methods into a simple yet effective algorithm. Other works propose different regularizations, \eg, enforcing a linear output~\cite{berthelot2019mixmatch} or minimizing the entropy~\cite{grandvalet2005semi}. 
However, none of these algorithms have been applied to the action detection problem. In this work, we try different types of SSL algorithms for action detection, including Mean Teacher~\cite{tarvainen2017mean}, FixMatch~\cite{sohn2020fixmatch}, and MixMatch~\cite{berthelot2019mixmatch}.

\textbf{Object-centric Action Understanding}.
Modern approaches for action recognition are built on top of deep models which take the whole frames as input to understand the action, \eg, 2D ConvNets~\cite{ioffe2015batch,he2016deep}, two-stream ConvNets~\cite{simonyan2014two,wang2016temporal}, and 3D ConvNets~\cite{tran2015learning,qiu2017learning,carreira2017quo}. Recent studies show that the action recognition models rely on the static appearance or the background scene, may degrade the recognition performance~\cite{zhou2018temporal} or lead to biased decisions~\cite{choi2019can}.
Since action is mainly characterized by the movement of foreground objects, we expect the deep models to focus on the foreground motion for better recognition.
Other works have also verified the superiority of object-centric representations~\cite{mccandless2013object,materzynska2020something,wang2018videos}. 
To this end, in order to better utilize unlabeled data in our task, we propose the UFA module to detect foreground. Previous approaches to foreground detection usually need large amount of labeled data~\cite{tokmakov2017learning,tokmakov2017learningb,song2018pyramid,cheng2017segflow}. A recent work learns the foreground detector based on the assumption that foreground motion and background motion in the same frame are mutually independent~\cite{yang2019unsupervised}, which may not hold for action videos, considering that background motion may be affected by foreground objects through camera motion. 
We take the camera motion into consideration and fix the drawback in previous assumption~\cite{yang2019unsupervised} by checking the neighboring frames.
%Our approach takes the camera motion into consideration and fixes the assumption by also looking at the neighboring frames.

\textbf{Weakly-supervised Action Detection}.
WSAD learns to predict the actionness score (the likelihood of a proposal being an action) with only video-level classification label. Popular WSAD methods can be categorized as top-down or bottom-up methods. Top-down methods~\cite{liu2019completeness,narayan20193c,paul2018w,wang2017untrimmednets} first train a video-level classifier and then obtain the proposal actionness score from the temporal class activation map (TCAM). Bottom-up methods~\cite{nguyen2018weakly,nguyen2019weakly,shou2018autoloc,yuan2018marginalized} directly predict the actionness score from raw proposals and learn to classify the video whose feature is given by averaging proposal features weighted by actionness score. In this work we adopt the bottom-up pipeline to train models on weakly-labeled data. However, it is known that WSAD methods are prone to recognize non-action frames as action (\ie action-context confusion) when non-action frames also contain the category-indicative information. To address this issue, Liu \etal~\cite{liu2019completeness} attempt to separate action and context with hard negative mining by assuming that context clips should be stationary. Shi \etal~\cite{shi2020weakly} separate action and context by modeling the feature-level distribution with a generative model. In this work, we design an information bottleneck to suppress the information in non-action frames.
%without additional modules or assumptions.
%while no additional modules or assumptions are required.

\section{Method}

\begin{figure*}[t]
\centering
\includegraphics[width=\linewidth]{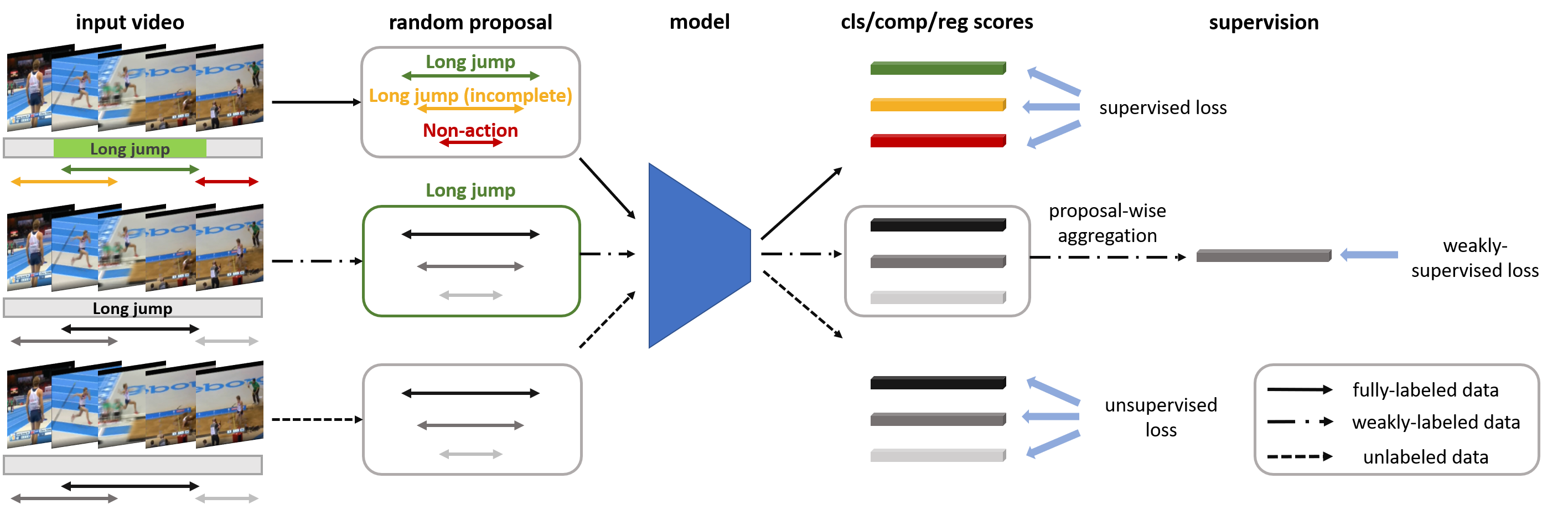}
   \caption{The basic pipeline of our proposed model for temporal action detection under multi-level supervision.}
\label{fig:pipeline}
% \vspace{-1em}
\end{figure*}

%Suppose we have a dataset of 
In temporal action detection, for a video $\mathbf{X}$ with $T$ frames (either RGB or optical flow) $\mathbf{X} = (\mathbf{x}_t)_{t = 1}^T$, we randomly sample $N$ proposals $(s_i, e_i)_{i = 1}^N$, where $s_i$ and $e_i$ are the start and end time of the $i$-th proposal. Normally, we utilize a trainable backbone $g_\theta$ to extract features for each frame, $\mathbf{z}_t = g_\theta(\mathbf{x}_t)$, and then obtain the feature $\mathbf{p}_i$ for each proposal by average pooling
\begin{equation}\small
    \label{eq:proposal_feature}
    \mathbf{p}_i = \frac{1}{e_i - s_i + 1} \sum_{t = s_i}^{e_i} \mathbf{z}_t.
\end{equation}
For fully-labeled data, each proposal $\mathbf{p}_i$ has a class label $y_i \in \{0, 1, \cdots, C\}$, where $C$ is the total number of action categories and $y_i = 0$ indicates a non-action proposal, and a regression score $r_i \in \mathbb{R}^2$ for start and end time. For weakly-labeled data, we only have the video-level class label $y \in \{1, \cdots, C\}$. 
No label is available for unlabeled data. We denote the subsets of labeled, weakly-labeled, and unlabeled data by $\mathcal{S}$, $\mathcal{W}$, and $\mathcal{U}$, respectively. 
The basic pipeline under multi-level supervision is shown in Fig.~\ref{fig:pipeline}.

%%%%%%%%%%%%%%%%%%%%%%%%%%%%%%%%%%%%%%%%%%%

\subsection{Semi-Supervised Action Detection Baselines}
\label{sec:method_semi_baseline}

%We first build the SSAD baselines by combining algorithms for Fully-Supervised Action Detection (FSAD) and Semi-Supervised Learning (SSL). 
We first build the SSAD baselines by integrating the Semi-Supervised Learning (SSL) algorithms into Fully-Supervised Action Detection (FSAD). We choose SSN~\cite{zhao2017temporal} as the basic FSAD method. For SSL, we implement the state-of-the-art algorithms including Mean Teacher~\cite{tarvainen2017mean}, MixMatch~\cite{berthelot2019mixmatch}, and FixMatch~\cite{sohn2020fixmatch} for our action detection setting.
We briefly recap the aforementioned FSAD and SSL algorithms in the remainder of this section.
%We introduce the FSAD and SSL algorithms separately in the remainder of this section.
% See Fig.~\ref{fig:semi_baseline} for an overall illustration of SSAD baselines.

In SSN, we train a classification module $h_{cls}$ and a regression module $h_{reg}$ at proposal level, and in the meantime we also train a completeness module $h_{comp}$ to predict the proposal completeness (denoted by $c_i \in \{0, 1\}$), indicating whether the proposal $\mathbf{p}_i$ is a complete action clip or not.%\footnote{Other tricks (\eg STPP, TAG) from the original SSN algorithm are removed for simplicity (Sec.~\ref{sec:implementation}).} 
A proposal is considered as incomplete ($c_i = 0$) if more than 80\% of its own span is overlapped with an action clip, while its IoU with the clip is below 0.3. The total loss $\mathcal{L}^S$ for the $i$-th proposal consists of three parts:
\begin{equation}\small
    \mathcal{L}^S(\mathbf{X}) = \mathcal{L}^S_{cls}(\mathbf{X}) + \alpha_{c}^S \mathcal{L}^S_{comp}(\mathbf{X}) + \alpha_{r}^S \mathcal{L}^S_{reg}(\mathbf{X}),
\end{equation}
where each loss is given by
\begin{equation}\small
\begin{split}
    \mathcal{L}^S_{cls} (\mathbf{X}) &= -\frac{1}{N} \sum_{i = 1}^N \log h_{cls}(y_i | \mathbf{p}_i), \\
    \mathcal{L}^S_{comp} (\mathbf{X}) &= -\frac{1}{N} \sum_{i = 1}^N \log h_{comp}(c_i | \mathbf{p}_i, y_i), \\
    \mathcal{L}^S_{reg} (\mathbf{X}) &= \frac{1}{N} \sum_{i = 1}^N \norm{h_{reg} (\mathbf{p}_i, c_i) - r_i}_1.
\end{split}
\end{equation}
%Note that we remove other tricks in the original SSN algorithm for simplicity (\eg STPP, TAG) (more details in Sec.~\ref{sec:implementation}).

%For SSL, we implement the state-of-the-art algorithms including Mean Teacher~\cite{tarvainen2017mean}, MixMatch~\cite{berthelot2019mixmatch}, and FixMatch~\cite{sohn2020fixmatch} for our action detection setting.

% %For SSL algorithms, each optimizes a manually designed loss on the unlabeled data:
% For the implementation of SSL algorithms (Mean Teacher~\cite{tarvainen2017mean}, MixMatch~\cite{berthelot2019mixmatch}, and FixMatch~\cite{sohn2020fixmatch}) based on the temporal action detection model SSN, we adapt the loss function $\mathcal{L}^U (\mathbf{X})$ of SSN for the unlabeled data considering the specific SSL algorithm (shown below).
For unlabeled data, we adapt the unsupervised loss of each SSL algorithm for action detection problem:
\begin{equation}\small
    \mathcal{L}^U (\mathbf{X}) = \mathcal{L}^U_{cls}(\mathbf{X}) + \alpha_{c}^U \mathcal{L}^U_{comp}(\mathbf{X}) + \alpha_{r}^U \mathcal{L}^U_{reg}(\mathbf{X}).
\end{equation}
In the next, we introduce the specific design of $\mathcal{L}_\ast^U (\mathbf{X})$ ($\ast$ $\in$ \{``\emph{cls}'', ``\emph{comp}'', ``\emph{reg}''\}) in each SSL algorithm.
%(Mean Teacher~\cite{tarvainen2017mean}, MixMatch~\cite{berthelot2019mixmatch}, and FixMatch~\cite{sohn2020fixmatch})

\textbf{Mean Teacher} optimizes the output consistency between two different augmentations of the same instance. It uses an Exponential Moving Average (EMA) of the backbone to extract features for one of the augmented input. The unsupervised loss is given~by
\begin{equation}\small
    \mathcal{L}_\ast^U (\mathbf{X}) = \frac{1}{N} \sum_{i = 1}^N \ell_\ast (h_\ast(\widetilde{\mathbf{p}}_i), h_\ast(\widetilde{\mathbf{p}}^{EMA}_i)),
\end{equation}
%where $\ast$ stands for \emph{cls}, \emph{comp} or \emph{reg},
where $h_\ast(\cdot)$ indicates the output of corresponding module, $\widetilde{\mathbf{p}}_i$ is the feature of the augmented proposal, and the \emph{EMA} superscript indicates the feature is extracted with the EMA version of the backbone. Note that $\widetilde{\mathbf{p}}_i$ and $\widetilde{\mathbf{p}}^{EMA}_i$ come from different inputs because the augmentation is stochastic. $\ell_\ast$ is the ``distance" in the corresponding output space. We use KL divergence for classification and completeness scores, and $L_1$ distance for regression scores.

\textbf{MixMatch} enforces a linear output between input points. 
%Since MixMatch works on unlabeled data, 
Following MixMatch method to deal with unlabeled data, we first obtain the pseudo label $\hat{h}_\ast (\mathbf{p}_i)$ for each proposal by sharpening the average output of $K$ randomly augmented inputs, and then train the model on the mixed pseudo-labeled data with
\begin{equation}\footnotesize
    \mathcal{L}_\ast^U (\mathbf{X}) = \frac{1}{N} \sum_{i = 1}^N \ell_\ast (h_\ast(\mathcal{M}_\lambda(\mathbf{p}_i, \mathbf{p}^\prime_i)), \mathcal{M}_\lambda(\hat{h}_\ast(\mathbf{p}_i), \hat{h}_\ast(\mathbf{p}^\prime_i))),
\end{equation}
where $\mathbf{p}^\prime_i$ is a proposal randomly sampled from the dataset, and $\mathcal{M}_\lambda(\cdot, \cdot)$ is the \emph{mixup} function which is basically a linear interpolation with weights $\lambda$ and $1 - \lambda$. $\lambda$ is sampled from a Beta distribution~\cite{berthelot2019mixmatch}. 

\textbf{FixMatch} combines consistency-based and pseudo-label methods. It takes the augmented input and trains the model to predict the pseudo label $\hat{h}_\ast (\mathbf{p}_i)$:
\begin{equation}\footnotesize
    \mathcal{L}_\ast^U (\mathbf{X}) = \frac{1}{N} \sum_{i = 1}^N \mathbbm{1} (\max(\hat{h}_\ast (\mathbf{p}_i)) > \tau) \ \cdot \ \ell_\ast (h_\ast(\widetilde{\mathbf{p}}_i), \hat{h}_\ast (\mathbf{p}_i)).
\end{equation}
The pseudo label is generated from the output of a \emph{weakly}-augmented input. 
Note that we only train on pseudo labels with high confidence, \ie, $\max(\hat{h}_\ast (\mathbf{p}_i)) > \tau$. Since this method does not apply to non-classification task such as regression, we do not add this term in $\mathcal{L}_{reg}^U$.

The overall objective for our proposed SSAD models is
\begin{equation}\small
    \mathcal{L} = \frac{1}{|\mathcal{S}|} \sum_\mathbf{X \in \mathcal{S}} \mathcal{L}^S (\mathbf{X}) + \alpha^U \cdot \frac{1}{|\mathcal{U}|} \sum_\mathbf{X \in \mathcal{U}} \mathcal{L}^U (\mathbf{X}).
\end{equation}

In all SSL algorithms, we exploit both spatial and temporal augmentations for video data. For spatial augmentations, we apply random noise and horizontal flip to all the frames in each proposal. We also design three temporal augmentations: (i) \textbf{Temporal Resampling}: In Eq.~\ref{eq:proposal_feature} we obtain the proposal feature by average pooling. In practice, we only sample $L$ frames from the proposal 
%by uniformly dividing the proposal into $L$ segments and randomly sampling one frame from each segment
, and take the average of their features as an efficient estimation~\cite{zhao2017temporal}. In Temporal Resampling, we resample $L$ frames from the proposal and take the new average as an augmented feature. (ii) \textbf{Temporal Resolution}: Instead of sampling $L$ frames from each proposal, we sample $2L$ or $L/2$ frames. (iii) \textbf{Temporal Flip}: The video is played backwards. For the weak augmentation in Fixmatch, we employ only spatial augmentations without temporal ones. Please refer to supplementary for an evaluation of the augmentations.

%%%%%%%%%%%%%%%%%%%%%%%%%%%%%%%%%%%%%%%%%%%

\subsection{Unsupervised Foreground Attention}

\begin{figure}[t]
\centering
\includegraphics[width=0.9\linewidth]{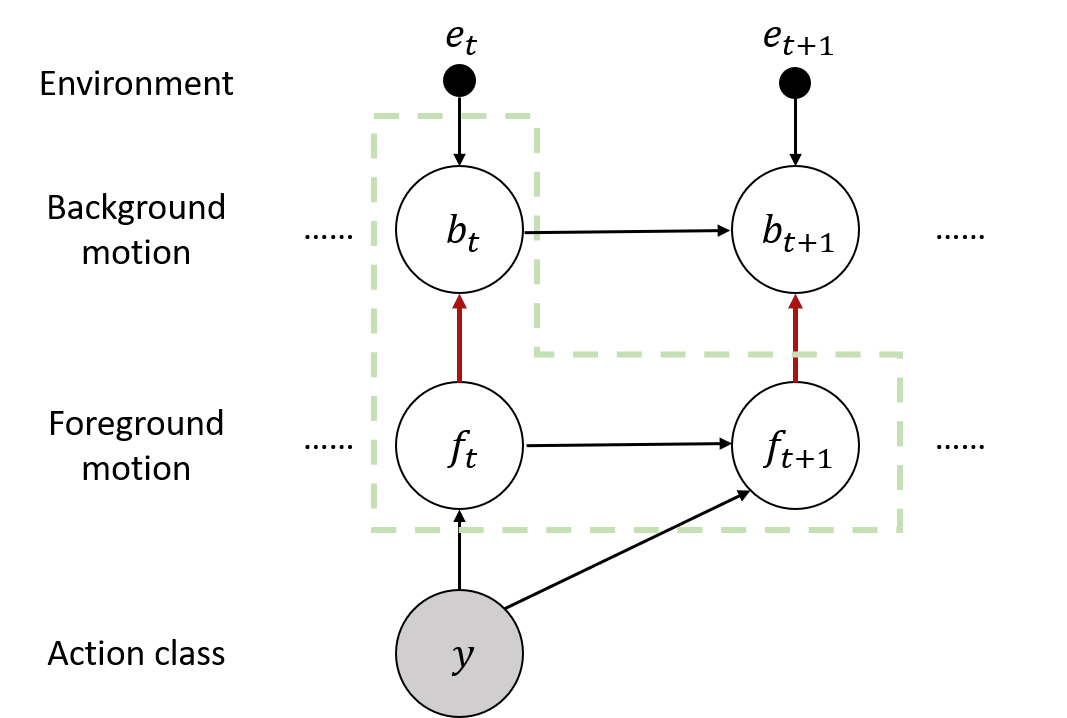}
   \caption{A (causal) graphical model for foreground/background motion in a video. Here we assume the motion is Markovian. The foreground motion $f_t$ is determined by the action category $y$. The background motion $b_t$ is affected by random factors $e_t$ in the environment. Meanwhile, $b_t$ is also affected by $f_t$ through camera motion (red arrows). The structure in the green box indicates the conditional independence $(b_t \indep f_{t+1}) | f_t$.}
\label{fig:unsupervised_attention}
\vspace{-1em}
\end{figure}

\textbf{The SSAD baseline is prone to miss part of the action (action incompleteness),} as revealed by the error analysis in Fig.~\ref{fig:intro}.
Since actions are basically defined by the movement of foreground objects, our conjecture is that the model can better recognize complete action clips by paying more attention to the foreground. Moreover, learning the attention without extra supervision can better utilize the unlabeled data. Therefore, we propose an unsupervised foreground attention (UFA) module to address the issue.

Intuitively, one can hardly estimate the motion of foreground from background, and vice versa, which indicates a certain level of ``independence'' between the \textbf{motions} of background and foreground.
%To separate foreground and background in each frame without supervision, we need to exploit the intrinsic structure of each frame. 
%Although background scene can be really informative of the foreground action (\eg we can tell it is golf playing by only seeing the golf course), there is a certain level of independence between the \textbf{motions} of background and foreground. 
\cite{yang2019unsupervised} assumes a complete independence between the foreground and background motions, which may not hold in action videos because of \textbf{camera motion}. 
%For example, the golf flag waving in the background is independent of the golf swing in the foreground. However, the foreground and background in a video are not totally independent because of \textbf{camera motion}. 
An intuitive example is that we usually move the camera to track the foreground object when recording a video, which consequently causes movement of background scene. This creates a case that background motion is partially affected by foreground motion.
%When we are recording video, we usually move our camera to follow the foreground object, which also makes the background move in this change of view. This is saying, the foreground motion can also causes background motion. 
Taking this into consideration, we describe the causal structure of foreground/background motions with the graph in Fig.~\ref{fig:unsupervised_attention}. We denote the foreground and background motions in frame $t$ by $f_t$ and $b_t$. $f_t$ is determined by the action category $y$, and $b_t$ is affected by random factors $e_t$ in the environment. $f_t$ has an effect on $b_t$ via camera motion. 
Although $f_t$ and $b_t$ are not completely independent, there is a conditional independence between $b_t$, $f_t$, and $f_{t+1}$, \ie, $I(b_t, f_{t+1} | f_t) = 0$, where $I$ is the (conditional) mutual information.
%We can find that the foreground the background are not independent. However, there is a conditional independence between $b_t$, $f_t$, and $f_{t+1}$, \ie, $I(b_t, f_{t+1} | f_t) = 0$, where $I$ is the (conditional) mutual information. 
Intuitively, this means the foreground motion is a self-contained flow which is not affected by the background.

Suppose $\hat{\mathbf{z}}_t \in \mathbb{R}^{h \times w \times c}$ is the feature map before the final spatial pooling, \ie, $\mathbf{z}_t = \textrm{Pool}(\hat{\mathbf{z}}_t) \in \mathbb{R}^c$. Now we apply the UFA module $\textrm{Att}_\phi(\cdot)$ on $\hat{\mathbf{z}}_t$, \ie, $\mathbf{z}_t = \textrm{Pool}(\mathbf{a}_t \ast \hat{\mathbf{z}}_t)$, where $\mathbf{a}_t = \textrm{Att}_\phi(\hat{\mathbf{z}}_t) \in \mathbb{R}^{h \times w \times 1}$ and $\ast$ is the broadcastable Hadamard product. To train UFA, we minimize the conditional mutual information:
\begin{equation}\small
\label{eq:objective_attention}
\begin{split}
    & \quad \quad\quad \min_\phi I(\mathbf{B}_t, \mathbf{F}_{t+1} | \mathbf{F}_t) \\
    \! \! \! \! \! \! \! \! \! \! \Leftrightarrow \quad &\min_\phi H(\mathbf{F}_{t+1} | \mathbf{F}_t) - H(\mathbf{F}_{t+1} | \mathbf{F}_t, \mathbf{B}_t) , 
\end{split}
\end{equation}
where $H$ is the (conditional) entropy, and $\mathbf{F}_t$ and $\mathbf{B}_t$ are the random variables for foreground and background \emph{motion} features. We denote their realizations by $\mathbf{f}_t$ and $\mathbf{b}_t$. When the input is optical flow, we directly extract motion feature from $\hat{\mathbf{z}}_t$, as the flow itself already represents motion:
\begin{equation}\small
\begin{split}
    \mathbf{f_t} &= \textrm{Pool}(\mathbf{a}_t \ast \hat{\mathbf{z}}_t), \\
    \mathbf{b_t} &= \textrm{Pool}((1 - \mathbf{a}_t) \ast \hat{\mathbf{z}}_t).
\end{split}
\end{equation}
When the input is RGB frame, we use the difference between two consecutive frame features as the motion feature
\begin{equation}\small
\begin{split}
    \mathbf{f_t} &= \textrm{Pool}(\mathbf{a}_t \ast (\hat{\mathbf{z}}_{t+1} - \hat{\mathbf{z}}_t)), \\
    \mathbf{b_t} &= \textrm{Pool}((1 - \mathbf{a}_t) \ast (\hat{\mathbf{z}}_{t+1} - \hat{\mathbf{z}}_t)).
\end{split}
\end{equation}
To optimize Eq.~\ref{eq:objective_attention}, we need an estimation of the entropy. If we assume the feature distributions are Gaussian, we have
\begin{equation}\small
\begin{split}
    H(\mathbf{F}_{t+1} | \mathbf{F}_t, \mathbf{B}_t) &\propto \log E_{(\mathbf{f}_t, \mathbf{b}_t, \mathbf{f}_{t+1})} \norm{\mathbf{f}_{t+1} - E(\mathbf{F}_{t+1} | \mathbf{f}_t, \mathbf{b}_t)}_2^2, \\
    H(\mathbf{F}_{t+1} | \mathbf{F}_t) &\propto \log E_{(\mathbf{f}_t, \mathbf{f}_{t+1})} \norm{\mathbf{f}_{t+1} - E(\mathbf{F}_{t+1} | \mathbf{f}_t)}_2^2.
\end{split}
\end{equation}
Therefore, we first train two predictors $u_\psi$ and $u_\zeta$ to approximate the conditional estimation:
\begin{equation}\small
\begin{split}
    \psi^\ast &= \argmin_\psi E_{(\mathbf{f}_t, \mathbf{b}_t, \mathbf{f}_{t+1})} \norm{\mathbf{f}_{t+1} - u_\psi(\mathbf{f}_t, \mathbf{b}_t)}_2^2, \\
    \zeta^\ast &= \argmin_\zeta E_{(\mathbf{f}_t, \mathbf{f}_{t+1})} \norm{\mathbf{f}_{t+1} - u_\zeta(\mathbf{f}_t)}_2^2,
\end{split}
\end{equation}
and then train UFA by optimizing
\begin{equation}\small
    \min_\phi \log \frac{E_{(\mathbf{f}_t, \mathbf{f}_{t+1})} \norm{\mathbf{f}_{t+1} - u_{\zeta^\ast}(\mathbf{f}_t)}_2^2}{ E_{(\mathbf{f}_t, \mathbf{b}_t, \mathbf{f}_{t+1})} \norm{\mathbf{f}_{t+1} - u_{\psi^\ast}(\mathbf{f}_t, \mathbf{b}_t)}_2^2}  .
\end{equation}
In practice, we update $\psi$, $\zeta$, and $\phi$ simultaneously to avoid the bi-level optimization. With our UFA module, we reduce the action incompleteness error and improve the performance of SSAD baselines for free (Sec.~\ref{sec:exp_ssal}).

%%%%%%%%%%%%%%%%%%%%%%%%%%%%%%%%%%%%%%%%%%%

\subsection{Omni-Supervised Action Detection with Information Bottleneck}

Now we add weakly-labeled data to train the model with three levels of supervision, and form Omni-Supervised Action Detection (OSAD). For fully labeled and unlabeled data, we minimize $\mathcal{L}^S(\mathbf{X})$ and $\mathcal{L}^U(\mathbf{X})$ as introduced in Sec.~\ref{sec:method_semi_baseline}. For weakly-labeled data, we optimize a video-level classification loss $\mathcal{L}^W(\mathbf{X})$ which is given by
\begin{equation}\small
    \mathcal{L}^W(\mathbf{X}) = -\log h_{cls} (y | \mathbf{X}),
\end{equation}
%where $h_{cls} (\cdot | \mathbf{X})$ is the video-level classification score given by the weighted average of the proposal classification scores
where $h_{cls} (y | \mathbf{X})$ is the video-level classification score given by the weighted average of the proposal classification scores
\begin{equation}\small
    h_{cls} (y | \mathbf{X}) =  \frac{\sum_{i = 1}^N \lambda_i h_{cls} (y | \mathbf{p}_i)}{\sum_{i = 1}^N \lambda_i},
\end{equation}
where $\lambda_i = 1 - h_{cls} (y=0 | \mathbf{p}_i)$ is the probability of $\mathbf{p}_i$ being an action. Then the overall loss for OSAD baseline is 
\begin{equation}\footnotesize
    \mathcal{L} = \frac{1}{|\mathcal{S}|} \sum_\mathbf{X \in \mathcal{S}} \mathcal{L}^S (\mathbf{X}) + \alpha^U \frac{1}{|\mathcal{U}|} \sum_\mathbf{X \in \mathcal{U}} \mathcal{L}^U (\mathbf{X}) + 
    \alpha^W \frac{1}{|\mathcal{W}|} \sum_\mathbf{X \in \mathcal{W}} \mathcal{L}^W (\mathbf{X}).
\end{equation}

However, \textbf{the OSAD baseline is prone to classify non-action frames as action frames (action-context confusion)}, as shown in Fig.~\ref{fig:intro}.
This issue is common when learning action detection from weakly-labeled data~\cite{liu2019completeness,shi2020weakly}. 
Ideally, the recognition model is expected to classify the weakly-labeled videos based on action information (\eg swimming) which could also benefit the action detection task.
%When training classification on weakly-labeled data, the ideal situation would be the model learning to classify based on action information (\eg swimming) which can also benefit localization during inference. 
However, the model tends to take a ``shortcut" instead and learn to classify action based on the scene information (\eg the swimming pool), which would disrupt the detection by mistaking non-action frames with scenes for action frames.
%which is not useful for discriminating action frames from non-action ones. 
Now the question is, \emph{how can we filter out the scene information and only keep the action information when only training the classification task}? Note that, although action frames contain both action and scene information, the non-action frames only contain the scene part. Thus, we propose to ``unlearn" the scene information by penalizing all the information extracted from non-action frames. 

Assuming the feature distribution to be Gaussian, the information of non-action frames can be estimated by
\begin{equation}\small
    I \propto E_{\mathbf{X}} \left\{ \frac{1}{\sum_i \Bar{\lambda}_i} \Bar{\lambda}_i \norm{\mathbf{p}_i - \frac{1}{\sum_i \Bar{\lambda}_i} \Bar{\lambda}_i \mathbf{p}_i}_2^2 \right\},
\end{equation}
where $\Bar{\lambda}_i = h_{cls}(y=0 | \mathbf{p}_i)$ is the likelihood of being a non-action proposal. Then we add this term into $\mathcal{L}^W$. Note that we also add a normalization term to avoid a trivial solution. The final loss on weakly-labeled data is
\begin{equation}\small
    \mathcal{L}^W(\mathbf{X}) = -\log h_{cls} (y | \mathbf{X}) + \alpha^W_I \frac{\frac{1}{\sum_i \Bar{\lambda}_i} \Bar{\lambda}_i \norm{\mathbf{p}_i - \frac{1}{\sum_i \Bar{\lambda}_i} \Bar{\lambda}_i \mathbf{p}_i}_2^2}{\frac{1}{\sum_i \Bar{\lambda}_i} \Bar{\lambda}_i \norm{\mathbf{p}_i}_2^2}.
\end{equation}
Intuitively, this is an explicit information bottleneck (IB)~\cite{tishby2000information}, where we maximize the (action) information about the classification label and meanwhile minimize the (scene) information about the non-action input frames. 
\section{Experiments}

\subsection{Datasets and Metrics}

We evaluate SSAD and OSAD models on two standard benchmarks, THUMOS14~\cite{idrees2017thumos} and ActivityNet1.2~\cite{caba2015activitynet}. 

\textbf{THUMOS14} consists of videos from 20 action classes. We follow the literature to train on validation set of 200 videos and evaluate on test set of 212 videos. 
This dataset has a fine level annotation of action. 
On average, each video lasts 3 minutes while containing 15.5 action clips. Duration of action instances varies from several seconds to minutes.
 
\textbf{ActivityNet1.2} contains \texttildelow10k videos from 100 classes. Each video has an average of 1.5 action clips. Following the literature, we train our model on training set of 4819 videos and evaluate on validation set of 2383 videos.

\textbf{Data Split}.
%Since both the datasets we use are fully-labeled, we need to first create labeled/unlabeled data splits for SSAD. 
Both datasets are originally used in fully-supervised action detection, and we need to create labeled/unlabeled data splits for our SSAD task. 
We create the data splits by randomly sampling fully-labeled data in a specific ratio, and treat the rest as unlabeled data. The same process is also applied for creating three disjoint splits in OSAD task. We also have a sanity check to show that the random sampling will not affect the performance significantly as long as the ratio of each split is fixed (see Sec.~\ref{sec:exp_ssal}).

\textbf{Evaluation Metrics.} Following the standard evaluation protocol, we report mean Average Precision (mAP) at different intersection over union (IoU) thresholds. We use \{0.3, 0.4, 0.5, 0.6, 0.7\} as the IoU thresholds for THUMOS14, and \{0.5, 0.75, 0.95\} for ActivityNet1.2. We also report the average mAP over thresholds [0.5 : 0.05 : 0.95] (mAP@AVG). 

%%%%%%%%%%%%%%%%%%%%%%%%%%%%%%%%%%%%%%%%%%%%%

\begin{table}[t]\small
\caption{\small Sanity check for the effect of random sampling in creating data splits for SSAD task. We randomly sample three data splits of ActivityNet1.2 in a fixed ratio of labeled/unlabeled data, and report the performance (mAP@AVG) of SSAD baselines on each split. }
\vspace{-0.45cm}
\label{table:sanity_check}
\begin{center}
\begin{tabular}{lccc}
\hline
 & split \#1 & split \#2 & split \#3 \\
\hline
MT & 11.49 & 11.41 & 11.56 \\
MixMatch & 11.05 & 11.03 & 11.08 \\
FixMatch & 11.88 & 11.89 & 11.97 \\
\hline
\end{tabular}
\end{center}
\vspace{-0.5cm}
\end{table}

\subsection{Implementation Details}
\label{sec:implementation}

\textbf{Input}. We use RGB and optical flow as two separate input streams. RGB frames are sampled at 25fps for THUMOS14 and 3fps for ActivityNet1.2. Then optical flow is extracted from RGB frames using TV-L1 algorithm~\cite{perez2013tv}. 
%, which is implemented in the DenseFlow toolbox\footnote{https://github.com/open-mmlab/denseflow}. 
During training, we use sliding window to generate proposals of various durations. Then we sample 5 frames from each proposal by uniformly dividing the proposal into 5 segments and randomly sampling one frame from each segment. Features are extracted from each frame and averaged as the proposal feature. Following SSN~\cite{zhao2017temporal}, we additionally sample 2 frames in [$s_i - \frac{e_i - s_i}{2}$, $s_i$] and 2 frames in [$e_i$, $e_i + \frac{e_i - s_i}{2}$] to include the temporal context when predicting completeness or regression scores. 

\textbf{Backbone Model}. We use BNInception~\cite{ioffe2015batch} as the classification backbone architecture in SSN, and replace the feature pyramid (STPP) in the original SSN algorithm with a simple temporal pooling. We also adopt sliding window instead of TAG for proposal sampling because TAG needs pretraining on labeled data which is partially unavailable in our setting. 

\textbf{Error Analysis}. We consider three common types of errors: 1) \textbf{Action incompleteness}: missing parts of action; 2) \textbf{Misclassification}: incorrect action classification; 3) \textbf{Action-context confusion}: recognizing non-action frames as action. Error analysis results are shown in Fig.~\ref{fig:intro} where Mixmatch is used in both SSAD and OSAD models.

\textbf{Hyperparameters}. Please refer to the supplementary.

%%%%%%%%%%%%%%%%%%%%%%%%%%%%%%%%%%%%%%%%%%%%%

\subsection{Semi-supervised Action Detection}
\label{sec:exp_ssal}

Before evaluating SSAD models, we first have a sanity check to show that randomly sampling in creating data splits will not greatly affect the result. We randomly sample 10\% labeled data from ActivityNet1.2 for three times and test the results of SSAD baselines with three SSL algorithms. As shown in Table~\ref{table:sanity_check}, the fluctuation in mAP@AVG is less than 0.1, so we experiment in one of the splits (split \#1).
%which means it is safe to run on one of the splits. 

\begin{table}[t]\small
\caption{\small Performance of SSAD models under 50\% supervision on THUMOS14. We report results of supervised-only (Sup) model, as well as SSAD baselines with Mean Teacher (MT), MixMatch and FixMatch, with or without UFA. We also list the results under 100\% supervision as reference.}
\label{table:ssal_thumos14}
\vspace{-0.45cm}
\begin{center}
\begin{tabular}{lcccccc}
\hline
Method & UFA & 0.3 & 0.4 & 0.5 & 0.6 & 0.7 \\
\hline
Sup (100\%) & - & 51.48 & 40.40 & 28.45 & 16.88 & 7.77 \\
\hline
% \multirow{7}{*}{10\%} & Supervised & - &  &  &  &  & \\
%  & \multirow{2}{*}{MT} & - &  &  &  &  & \\
%  &  & \cmark &  &  &  &  & \\
%  & \multirow{2}{*}{Mixmatch} & - &  &  &  &  & \\
%  &  & \cmark &  &  &  &  & \\
%  & \multirow{2}{*}{Fixmatch} & - &  &  &  &  & \\
%  &  & \cmark &  &  &  &  & \\
% \hline
Sup & - & 45.37 & 35.03 & 24.74 & 14.66 & 6.45 \\
\multirow{2}{*}{MT} & - & 43.74 & 34.62 & 24.62 & 14.90 & 6.87 \\
  & \cmark & 45.47 & 35.52 & 25.43 & 15.35 & 7.08 \\
 \multirow{2}{*}{MixMatch} & - & 44.79 & 35.80 & 24.79 & 14.76 & 6.60 \\
  & \cmark & \textbf{45.65} & \textbf{36.43} & \textbf{26.18} & \textbf{15.52} & \textbf{7.10} \\
 \multirow{2}{*}{FixMatch} & - & 42.94 & 33.19 & 23.58 & 14.10 & 6.61 \\
  & \cmark & 42.99 & 34.65 & 24.68 & 14.72 & 6.61 \\
\hline
\end{tabular}
\end{center}
\vspace{-0.45cm}
\end{table}

\begin{table}[t]\small
\caption{\small Performance of SSAD models under 10\% supervision on ActivityNet1.2 (the same setting to~THUMOS14). 
%We report results of supervised-only (Sup) model, as well as SSAD baselines with Mean Teacher (MT), MixMatch and FixMatch, with or without UFA. We also list the results under 100\% supervision as reference.
}
\label{table:ssal_anet}
\vspace{-0.45cm}
\begin{center}
\begin{tabular}{lcccccc}
\hline
Method & UFA & 0.5 & 0.75 & 0.95 & AVG \\
\hline
Sup (100\%) & - & 32.88 & 18.84 & 3.03 & 19.44 \\
\hline
Sup & - & 18.30 & 11.47 & 1.61 & 11.29 \\
\multirow{2}{*}{MT} & - & 18.71 & 11.55 & 1.55 & 11.49 \\
 & \cmark & 19.11 & 11.75 & 1.63 & 11.74 \\
\multirow{2}{*}{MixMatch} & - & 17.84 & 11.27 & 1.61 & 11.05 \\
 & \cmark & 18.73 & 11.91 & 1.82 & 11.84 \\
\multirow{2}{*}{FixMatch} & - & 18.63 & 12.03 & 1.68 & 11.88 \\
 & \cmark & \textbf{19.47} & \textbf{12.54} & \textbf{1.88} & \textbf{12.27} \\
\hline
% \multirow{7}{*}{50\%} & Supervised & - &  &  &  & \\
%  & \multirow{2}{*}{MT} & - &  &  &  & \\
%  &  & \cmark &  &  &  & \\
%  & \multirow{2}{*}{Mixmatch} & - &  &  &  & \\
%  &  & \cmark &  &  &  & \\
%  & \multirow{2}{*}{Fixmatch} & - &  &  &  & \\
%  &  & \cmark &  &  &  & \\
% \hline
\end{tabular}
\end{center}
\vspace{-0.55cm}
\end{table}

Table~\ref{table:ssal_thumos14} shows the performance of SSAD models on THUMOS14. Since each class only has 10 videos in THUMOS14, it will be more like a ``few-shot" setting (1 video per class) if we choose the split of 10\% labeled data which is normally adopted in SSL community. Thus, we test on the data split of 50\% / 50\% (labeled / unlabeled). We also report the result with 100\% labeled data for reference. As we can see, the SSAD baselines barely improve the performance over supervised-only model due to the action incompleteness issue (Fig.~\ref{fig:intro}).
The FixMatch method even slightly degrades the precision. When adding the  UFA module, we improve the SSAD baselines by a large margin and fill in the result gap between 50\% supervised and 100\% supervised models with no extra labels.

SSAD results on ActivityNet1.2 are shown in Table~\ref{table:ssal_anet}. We test on the data split of 10\% / 90\% (labeled / unlabeled) as a usual practice in SSL. Similar to THUMOS14, we observe no improvement in Mean Teacher and MixMatch baselines, and an improvement of 0.5\% in FixMatch baseline. 
Our proposed unsupervised foreground attention brings an accuracy boost of 0.5\%-1\% on average for all the three SSAD baselines.
%The foreground attention brings a further boost of 0.5\%-1\% accuracy.

\begin{table}[t]\small
\caption{\small Comparison of different attention schemes. Here we report the performance of SSAD models with Mixmatch under 50\% supervision on THUMOS14. }
\label{table:attention_comparison}
\vspace{-0.45cm}
\begin{center}
\begin{tabular}{lccccc}
\hline
 & 0.3 & 0.4 & 0.5 & 0.6 & 0.7 \\
\hline
w/o att & 44.79 & 35.80 & 24.79 & 14.76 & 6.60 \\
Gaussian & 45.04 & 35.78 & 24.59 & 14.84 & 6.55 \\
CIS & 44.88 & 35.85 & 25.01 & 15.04 & 6.46 \\
UFA (ours) & \textbf{45.65} & \textbf{36.43} & \textbf{26.18} & \textbf{15.52} & \textbf{7.10} \\
\hline
\end{tabular}
\end{center}
\vspace{-0.25cm}
\end{table}

\begin{figure}[t]
\begin{center}
   \includegraphics[width=1\linewidth]{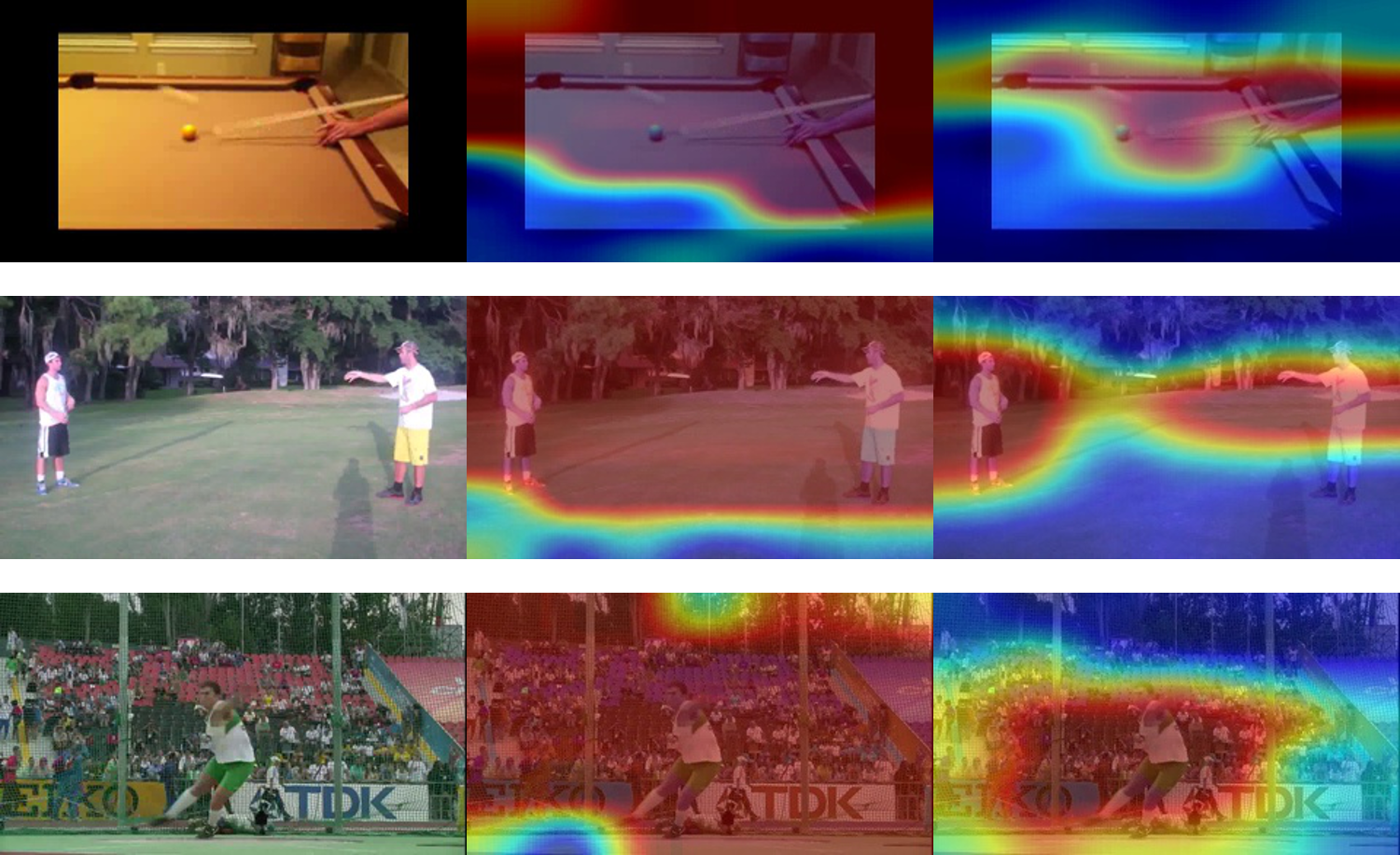}
\end{center}
\vspace{-0.45cm}
   \caption{\small Visualization of the foreground attention from three example videos (One frame per video is shown). Columns from left to right: original frame, CIS attention, and our proposed UFA.}
\label{fig:att}
\vspace{-0.2cm}
\end{figure}

%To further evaluate 
We further ablate the proposed UFA module, and compare it with a simple Gaussian attention and a state-of-the-art unsupervised foreground detection method called CIS~\cite{yang2019unsupervised}. The Gaussian attention is a simple baseline which utilizes a Gaussian distribution centered in the image as a fixed attention map. CIS learns to detect foreground based on the assumption of independence between foreground and background motion in the same frame, which neglects the factor of camera motion. As shown in Table~\ref{table:attention_comparison}, Gaussian and CIS have slight or even no improvement over the baseline without attention. For an intuitive understanding, we also visualize the attention map from both CIS and our UFA in Fig~\ref{fig:att}. 
As expected, our UFA module exhibits more clear foreground attention while suppresses the background, and further helps recognize complete action.
%As we expected, our proposed module more clearly detects the foreground object while represses the background scene. 
%This means focusing on the foreground indeed helps recognizing complete actions.

%%%%%%%%%%%%%%%%%%%%%%%%%%%%%%%%%%%%%%%%%%%%%

\subsection{Omni-Supervised Action Detection}

% \begin{table}[t]\small
% \caption{\small OSAD results on THUMOS14. We ablate on fully-supervised, unsupervised and weakly-supervised loss, as well as the proposed information bottleneck (IB). We also report the 100\% fully-supervised performance for reference. }
% \label{table:osal_thumos14}
% \begin{center}
% \begin{tabular}{c cccc c}
% \hline
% $|\mathcal{S}|: |\mathcal{W}|: |\mathcal{U}|$ & $\mathcal{L}^S$ & $\mathcal{L}^U$ & $\mathcal{L}^W$ & IB & mAP@0.5 \\
% \hline
% Full & \cmark & - & - & - & 28.45 \\
% \hline
% \multirow{4}{*}{0.1: 0.2: 0.7} & \cmark & - & - & - &  8.08 \\
%  & \cmark & \cmark & - & - & 8.96 \\
%  & \cmark & \cmark & \cmark & - & 11.37 \\
%  & \cmark & \cmark & \cmark & \cmark & \textbf{12.97} \\
% \hline
% \multirow{4}{*}{0.1: 0.4: 0.5} & \cmark & - & - & - & 8.08 \\
%  & \cmark & \cmark & - & - & 8.96 \\
%  & \cmark & \cmark & \cmark & - & 14.26 \\
%  & \cmark & \cmark & \cmark & \cmark & \textbf{15.33} \\
% \hline
% \end{tabular}
% \end{center}
% \end{table}

\begin{table}[t]\small
\caption{\small OSAD results on THUMOS14. We evaluate on two sets of data splits, 10\% / 20\% / 70\% and 10\% / 40\% / 50\% (fully-labeled / weakly-labeled / unlabeled). 
We report performances of supervised-only, semi-supervised model (SSAD) and omni-supervised models (OSAD and OSAD-IB). We also report the 100\% fully-supervised performance for reference. }
\label{table:osal_thumos14}
\vspace{-0.45cm}
\begin{center}
\begin{tabular}{c c c c}
\hline
Data split & Supervision & mAP@0.5 \\
\hline
100\% / 0\% / 0\% & Sup. only & 28.45 \\
\hline
\multirow{4}{*}{10\% / 20\% / 70\%} & Sup. only &  8.08 \\
 & SSAD & 8.96 \\
 & OSAD & 11.37 \\
 & OSAD-IB & \textbf{12.97} \\
\hline
\multirow{4}{*}{10\% / 40\% / 50\%} & Sup. only & 8.08 \\
 & SSAD & 8.96 \\
 & OSAD & 14.26 \\
 & OSAD-IB & \textbf{15.33} \\
\hline
\end{tabular}
\end{center}
\vspace{-0.45cm}
\end{table}

% \begin{table}[t]\small
% \caption{\small OSAD results on ActivityNet1.2.}
% \label{table:osal_anet}
% \begin{center}
% \begin{tabular}{c cccc c}
% \hline
% $|\mathcal{S}|: |\mathcal{W}|: |\mathcal{U}|$ & $\mathcal{L}^S$ & $\mathcal{L}^U$ & $\mathcal{L}^W$ & IB & mAP@AVG \\
% \hline
% Full & \cmark & - & - & - & 19.44 \\
% \hline
% \multirow{4}{*}{0.1: 0.2: 0.7} & \cmark & - & - & - &  11.29 \\
%  & \cmark & \cmark & - & - & 11.84 \\
%  & \cmark & \cmark & \cmark & - & 11.78 \\
%  & \cmark & \cmark & \cmark & \cmark & \textbf{12.30} \\
% \hline
% \multirow{4}{*}{0.1: 0.4: 0.5} & \cmark & - & - & - &  11.29 \\
%  & \cmark & \cmark & - & - & 11.84 \\
%  & \cmark & \cmark & \cmark & - & 12.44 \\
%  & \cmark & \cmark & \cmark & \cmark & \textbf{13.25} \\
% \hline
% \end{tabular}
% \end{center}
% \end{table}

We evaluate the OSAD models on two sets of data splits, 10\% / 20\% / 70\% and 10\% / 40\% / 50\% (fully-labeled / weakly-labeled / unlabeled), on both datasets. 
In all experiments, we unify the experimental setting of using Mixmatch on unlabeled data with UFA.
%In all experiments we use Mixmatch on unlabeled data and also apply UFA.
Table~\ref{table:osal_thumos14} shows the OSAD results on THUMOS14. We observe that by adding weakly-labeled data, the OSAD baseline brings a significant result improvement over the SSAD model. 
The final model OSAD-IB with our proposed information bottleneck further boosts the accuracy.
%The proposed IB further boosts the accuracy.

The OSAD results on ActivityNet1.2 are shown in Table~\ref{table:osal_anet}. Notably, introducing weakly-labeled data in the OSAD baseline only has a small impact on mAP@AVG compared to the SSAD model, due to the raised action-context confusion issue. After applying the proposed IB in our full OSAD-IB model, the action-context confusion issue is alleviated and a better performance is obtained.
%we reduce the Type-3 errors and get a better performance.

%%%%%%%%%%%%%%%%%%%%%%%%%%%%%%%%%%%%%%%%%%%%%

\subsection{Multi-Level Supervision is More Efficient}

%In practice, when confronting a specific application scenario, we first need to collect and annotate some training data from the same data source. 
%when we need to train an action detection model for a specific scenario, 
Practically, when confronting a specific application scenario, as a first step we need to collect and annotate some training data. 
%Normally, we have a budget of annotation cost which permits only a limited amount of labeled data. Then choosing the best annotation policy is particularly important. 
Suppose we have a fixed annotation budget which allows annotating limited amount of data, then designing the best annotation policy to maximize the detection performance is particularly useful considering the trade-off of supervision strength and annotation cost. 
To be more specific, we need to decide whether to annotate more less-expensive weak data, or less more-expensive data with full supervision, or adopt a mixed strategy. In this section, we simulate this scenario on THUMOS14 to explore and compare different annotation strategies, and show the benefit of our OSAD-IB model with multi-level supervision. 
%To be more specific, we need to decide whether to annotate more data with weak supervision, or less data with full supervision, or adopt a mixed strategy. In this section, we simulate this scenario on THUMOS14 to explore and compare different annotation strategies. 

% We first estimate the annotation cost for full and weak supervision of THUMOS14. THUMOS14 contains only 200 training videos collected from Youtube, which may be too few to obtain an accurate estimation of the annotation cost. Therefore, we collect similar videos from Youtube in a larger number of \texttildelow70,000, and record the annotation time. Specifically, it costs 312h to fully annotate 8,000 videos and 298h to weakly annotate 60,000 videos. Then the annotation cost for full and weak supervision are $140.4 \  \text{sec/vid}$ and $17.88 \ \text{sec/vid}$ separately, with the ratio of about $8: 1$.

\begin{table}[t]\small
\caption{\small OSAD results on ActivityNet1.2 (the same setting to~THUMOS14).}
\label{table:osal_anet}
\vspace{-0.45cm}
\begin{center}
\begin{tabular}{c c c c}
\hline
Data split & Method & mAP@AVG \\
\hline
100\% / 0\% / 0\% & Sup. only & 19.44 \\
\hline
\multirow{4}{*}{10\% / 20\% / 70\%} & Sup. only &  11.29 \\
 & SSAD & 11.84 \\
 & OSAD & 11.78 \\
 & OSAD-IB & \textbf{12.30} \\
\hline
\multirow{4}{*}{10\% / 40\% / 50\%} & Sup. only &  11.29 \\
 & SSAD & 11.84 \\
 & OSAD & 12.44 \\
 & OSAD-IB & \textbf{13.25} \\
\hline
\end{tabular}
\end{center}
\vspace{-0.45cm}
\end{table}

\begin{table}[t]\small
\caption{\small Different annotation policies under a fixed budget on THUMOS14. All the models use Mixmatch with UFA and IB. }
\label{table:annotation_budget}
\vspace{-0.45cm}
\begin{center}
\resizebox{1\columnwidth}{!}{
\begin{tabular}{lcccccc}
\hline
Policy & $|\mathcal{S}|$ (\%) & $|\mathcal{W}|$ (\%) & $|\mathcal{U}|$ (\%) & 0.3 & 0.5 & 0.7 \\
\hline
Full & 20\% & 0\% & 80\% & 34.02 & 16.06 & \textbf{3.92}  \\
\multirow{2}{*}{Mixed} & 15\% & 40\% & 45\% & \textbf{37.08} & \textbf{16.71} & 3.68\\
 & 10\% & 80\% & 20\% & 34.39 & 15.26 & 3.76 \\
Weak  & 8\% & 92\% & 0\% & 33.80 & 15.37 & 3.32  \\
\hline
\end{tabular}}
\end{center}
\vspace{-0.5cm}
\end{table}

As estimated from our user studies (see supplementary), the ratio of full and weak annotation cost on THUMOS14 is about $8: 1$. If we assume the weak annotation costs 1 time unit for each video, then the cost of fully annotating all 200 videos in THUMOS14 would be $1600 \ \text{units}$. We assume only a 20\% budget is available, \ie, $1600 \times 0.2 = 320 \ \text{units}$. We test three annotation policies: 1) \textbf{Full}: use all the budget on full supervision; 2) \textbf{Weak}: use all the budget on weak supervision, then on full supervision if there is any left; 3) \textbf{Mixed}: trade-off between Full and Weak. From Table~\ref{table:annotation_budget}, we observe that the Weak policy gives the worst result, which means full supervision is very important. However, spending all the budget on full supervision is also suboptimal in terms of results, and the best strategy is mixing full and weak supervision. The performance reaches a peak at $|\mathcal{S}|: |\mathcal{W}| = 15\% : 40\%$, suggesting that multi-level supervision is more efficient than the common strategy with only full or weak supervision. 
\section{Conclusion}

In this work we explore the utilization of data with multi-level supervision in temporal action detection. 
%unlabeled data in action detection.
We first introduce the semi-supervised action detection (SSAD) task to learn with both fully-labeled and unlabeled videos. 
%We first visit the task of semi-supervised action detection (SSAD) to train with both fully-labeled and unlabeled videos. 
We build the SSAD baselines by combining the fully-supervised action detection backbone with state-of-the-art semi-supervised learning algorithms. An unsupervised foreground attention (UFA) module is proposed to alleviate the action incompleteness issue in SSAD baselines by extracting object-centric features.
%extract object-centric representations and alleviate the action incompleteness problem in SSAD baselines. 
Then we study the task of omni-supervised action detection (OSAD) where weakly-labeled videos are further incorporated to learn a model with three levels of supervision.
%we further add weakly-labeled videos to train the model with three levels of supervision. 
To tackle the action-context confusion issue in OSAD baselines, we design an information bottleneck (IB) method to filter out scene information while keeping the action information so that the model can better distinguish action from context frames. 
We conduct extensive experiments on SSAD and OSAD baselines, as well as the proposed UFA and IB methods, showing their effectiveness. 
We further show the advantage of multi-level supervision over single supervision under a realistic scenario with a fixed annotation budget.

{\small
\bibliographystyle{ieee_fullname}
\bibliography{egbib}

\begin{thebibliography}{10}\itemsep=-1pt

\bibitem{berthelot2019mixmatch}
David Berthelot, Nicholas Carlini, Ian Goodfellow, Nicolas Papernot, Avital
  Oliver, and Colin~A Raffel.
\newblock Mixmatch: A holistic approach to semi-supervised learning.
\newblock In {\em Advances in Neural Information Processing Systems}, pages
  5049--5059, 2019.

\bibitem{caba2015activitynet}
Fabian Caba~Heilbron, Victor Escorcia, Bernard Ghanem, and Juan Carlos~Niebles.
\newblock Activitynet: A large-scale video benchmark for human activity
  understanding.
\newblock In {\em Proceedings of the IEEE Conference on Computer Vision and
  Pattern Recognition (CVPR)}, pages 961--970, 2015.

\bibitem{carreira2017quo}
Joao Carreira and Andrew Zisserman.
\newblock Quo vadis, action recognition? a new model and the kinetics dataset.
\newblock In {\em Proceedings of the IEEE Conference on Computer Vision and
  Pattern Recognition (CVPR)}, pages 6299--6308, 2017.

\bibitem{chao2018rethinking}
Yu-Wei Chao, Sudheendra Vijayanarasimhan, Bryan Seybold, David~A Ross, Jia
  Deng, and Rahul Sukthankar.
\newblock Rethinking the faster r-cnn architecture for temporal action
  localization.
\newblock In {\em Proceedings of the IEEE Conference on Computer Vision and
  Pattern Recognition}, pages 1130--1139, 2018.

\bibitem{cheng2017segflow}
Jingchun Cheng, Yi-Hsuan Tsai, Shengjin Wang, and Ming-Hsuan Yang.
\newblock Segflow: Joint learning for video object segmentation and optical
  flow.
\newblock In {\em Proceedings of the IEEE international conference on computer
  vision}, pages 686--695, 2017.

\bibitem{choi2019can}
Jinwoo Choi, Chen Gao, Joseph~CE Messou, and Jia-Bin Huang.
\newblock Why can't i dance in the mall? learning to mitigate scene bias in
  action recognition.
\newblock In {\em Advances in Neural Information Processing Systems}, pages
  853--865, 2019.

\bibitem{fathi2013modeling}
Alireza Fathi and James~M Rehg.
\newblock Modeling actions through state changes.
\newblock In {\em Proceedings of the IEEE Conference on Computer Vision and
  Pattern Recognition}, pages 2579--2586, 2013.

\bibitem{girshick2015fast}
Ross Girshick.
\newblock Fast r-cnn.
\newblock In {\em Proceedings of the IEEE International Conference on Computer
  Vision (ICCV)}, pages 1440--1448, 2015.

\bibitem{girshick2014rich}
Ross Girshick, Jeff Donahue, Trevor Darrell, and Jitendra Malik.
\newblock Rich feature hierarchies for accurate object detection and semantic
  segmentation.
\newblock In {\em Proceedings of the IEEE Conference on Computer Vision and
  Pattern Recognition (CVPR)}, pages 580--587, 2014.

\bibitem{grandvalet2005semi}
Yves Grandvalet and Yoshua Bengio.
\newblock Semi-supervised learning by entropy minimization.
\newblock In {\em Advances in neural information processing systems}, pages
  529--536, 2005.

\bibitem{he2016deep}
Kaiming He, Xiangyu Zhang, Shaoqing Ren, and Jian Sun.
\newblock Deep residual learning for image recognition.
\newblock In {\em Proceedings of the IEEE conference on computer vision and
  pattern recognition}, pages 770--778, 2016.

\bibitem{idrees2017thumos}
Haroon Idrees, Amir~R Zamir, Yu-Gang Jiang, Alex Gorban, Ivan Laptev, Rahul
  Sukthankar, and Mubarak Shah.
\newblock The thumos challenge on action recognition for videos “in the
  wild”.
\newblock {\em Computer Vision and Image Understanding (CVIU)}, 155:1--23,
  2017.

\bibitem{ioffe2015batch}
Sergey Ioffe and Christian Szegedy.
\newblock Batch normalization: Accelerating deep network training by reducing
  internal covariate shift.
\newblock {\em arXiv preprint arXiv:1502.03167}, 2015.

\bibitem{laine2016temporal}
Samuli Laine and Timo Aila.
\newblock Temporal ensembling for semi-supervised learning.
\newblock {\em arXiv preprint arXiv:1610.02242}, 2016.

\bibitem{lee2013pseudo}
Dong-Hyun Lee.
\newblock Pseudo-label: The simple and efficient semi-supervised learning
  method for deep neural networks.
\newblock In {\em Workshop on challenges in representation learning, ICML},
  volume~3, 2013.

\bibitem{lin2018bsn}
Tianwei Lin, Xu Zhao, Haisheng Su, Chongjing Wang, and Ming Yang.
\newblock Bsn: Boundary sensitive network for temporal action proposal
  generation.
\newblock In {\em Proceedings of the European Conference on Computer Vision
  (ECCV)}, pages 3--19, 2018.

\bibitem{liu2019completeness}
Daochang Liu, Tingting Jiang, and Yizhou Wang.
\newblock Completeness modeling and context separation for weakly supervised
  temporal action localization.
\newblock In {\em Proceedings of the IEEE Conference on Computer Vision and
  Pattern Recognition}, pages 1298--1307, 2019.

\bibitem{liu2016ssd}
Wei Liu, Dragomir Anguelov, Dumitru Erhan, Christian Szegedy, Scott Reed,
  Cheng-Yang Fu, and Alexander~C Berg.
\newblock Ssd: Single shot multibox detector.
\newblock In {\em European Conference on Computer Vision (ECCV)}, pages 21--37.
  Springer, 2016.

\bibitem{luo2020weakly}
Zhekun Luo, Devin Guillory, Baifeng Shi, Wei Ke, Fang Wan, Trevor Darrell, and
  Huijuan Xu.
\newblock Weakly-supervised action localization with expectation-maximization
  multi-instance learning.
\newblock {\em arXiv preprint arXiv:2004.00163}, 2020.

\bibitem{materzynska2020something}
Joanna Materzynska, Tete Xiao, Roei Herzig, Huijuan Xu, Xiaolong Wang, and
  Trevor Darrell.
\newblock Something-else: Compositional action recognition with
  spatial-temporal interaction networks.
\newblock In {\em Proceedings of the IEEE/CVF Conference on Computer Vision and
  Pattern Recognition}, pages 1049--1059, 2020.

\bibitem{mccandless2013object}
Tomas McCandless and Kristen Grauman.
\newblock Object-centric spatio-temporal pyramids for egocentric activity
  recognition.
\newblock In {\em BMVC}, volume~2, page~3. Citeseer, 2013.

\bibitem{narayan20193c}
Sanath Narayan, Hisham Cholakkal, Fahad~Shahbaz Khan, and Ling Shao.
\newblock 3c-net: Category count and center loss for weakly-supervised action
  localization.
\newblock In {\em Proceedings of the IEEE International Conference on Computer
  Vision}, pages 8679--8687, 2019.

\bibitem{nguyen2018weakly}
Phuc Nguyen, Ting Liu, Gautam Prasad, and Bohyung Han.
\newblock Weakly supervised action localization by sparse temporal pooling
  network.
\newblock In {\em Proceedings of the IEEE Conference on Computer Vision and
  Pattern Recognition}, pages 6752--6761, 2018.

\bibitem{nguyen2019weakly}
Phuc~Xuan Nguyen, Deva Ramanan, and Charless~C Fowlkes.
\newblock Weakly-supervised action localization with background modeling.
\newblock In {\em Proceedings of the IEEE International Conference on Computer
  Vision}, pages 5502--5511, 2019.

\bibitem{oliver2018realistic}
Avital Oliver, Augustus Odena, Colin~A Raffel, Ekin~Dogus Cubuk, and Ian
  Goodfellow.
\newblock Realistic evaluation of deep semi-supervised learning algorithms.
\newblock In {\em Advances in Neural Information Processing Systems}, pages
  3235--3246, 2018.

\bibitem{paul2018w}
Sujoy Paul, Sourya Roy, and Amit~K Roy-Chowdhury.
\newblock W-talc: Weakly-supervised temporal activity localization and
  classification.
\newblock In {\em Proceedings of the European Conference on Computer Vision
  (ECCV)}, pages 563--579, 2018.

\bibitem{perez2013tv}
Javier~S{\'a}nchez P{\'e}rez, Enric Meinhardt-Llopis, and Gabriele Facciolo.
\newblock Tv-l1 optical flow estimation.
\newblock {\em Image Processing On Line (IPOL)}, 2013:137--150, 2013.

\bibitem{qiu2017learning}
Zhaofan Qiu, Ting Yao, and Tao Mei.
\newblock Learning spatio-temporal representation with pseudo-3d residual
  networks.
\newblock In {\em Proceedings of the IEEE International Conference on Computer
  Vision (ICCV)}, pages 5533--5541, 2017.

\bibitem{redmon2016you}
Joseph Redmon, Santosh Divvala, Ross Girshick, and Ali Farhadi.
\newblock You only look once: Unified, real-time object detection.
\newblock In {\em Proceedings of the IEEE Conference on Computer Vision and
  Pattern Recognition (CVPR)}, pages 779--788, 2016.

\bibitem{ren2015faster}
Shaoqing Ren, Kaiming He, Ross Girshick, and Jian Sun.
\newblock Faster r-cnn: Towards real-time object detection with region proposal
  networks.
\newblock In {\em Advances in neural information processing systems (NeurIPS)},
  pages 91--99, 2015.

\bibitem{shi2020weakly}
Baifeng Shi, Qi Dai, Yadong Mu, and Jingdong Wang.
\newblock Weakly-supervised action localization by generative attention
  modeling.
\newblock In {\em Proceedings of the IEEE/CVF Conference on Computer Vision and
  Pattern Recognition}, pages 1009--1019, 2020.

\bibitem{shou2017cdc}
Zheng Shou, Jonathan Chan, Alireza Zareian, Kazuyuki Miyazawa, and Shih-Fu
  Chang.
\newblock Cdc: Convolutional-de-convolutional networks for precise temporal
  action localization in untrimmed videos.
\newblock In {\em Proceedings of the IEEE conference on computer vision and
  pattern recognition}, pages 5734--5743, 2017.

\bibitem{shou2018autoloc}
Zheng Shou, Hang Gao, Lei Zhang, Kazuyuki Miyazawa, and Shih-Fu Chang.
\newblock Autoloc: Weakly-supervised temporal action localization in untrimmed
  videos.
\newblock In {\em Proceedings of the European Conference on Computer Vision
  (ECCV)}, pages 154--171, 2018.

\bibitem{shou2016temporal}
Zheng Shou, Dongang Wang, and Shih-Fu Chang.
\newblock Temporal action localization in untrimmed videos via multi-stage
  cnns.
\newblock In {\em Proceedings of the IEEE Conference on Computer Vision and
  Pattern Recognition (CVPR)}, pages 1049--1058, 2016.

\bibitem{simonyan2014two}
Karen Simonyan and Andrew Zisserman.
\newblock Two-stream convolutional networks for action recognition in videos.
\newblock In {\em Advances in neural information processing systems}, pages
  568--576, 2014.

\bibitem{sohn2020fixmatch}
Kihyuk Sohn, David Berthelot, Chun-Liang Li, Zizhao Zhang, Nicholas Carlini,
  Ekin~D Cubuk, Alex Kurakin, Han Zhang, and Colin Raffel.
\newblock Fixmatch: Simplifying semi-supervised learning with consistency and
  confidence.
\newblock {\em arXiv preprint arXiv:2001.07685}, 2020.

\bibitem{song2018pyramid}
Hongmei Song, Wenguan Wang, Sanyuan Zhao, Jianbing Shen, and Kin-Man Lam.
\newblock Pyramid dilated deeper convlstm for video salient object detection.
\newblock In {\em Proceedings of the European conference on computer vision
  (ECCV)}, pages 715--731, 2018.

\bibitem{tarvainen2017mean}
Antti Tarvainen and Harri Valpola.
\newblock Mean teachers are better role models: Weight-averaged consistency
  targets improve semi-supervised deep learning results.
\newblock In {\em Advances in neural information processing systems}, pages
  1195--1204, 2017.

\bibitem{tishby2000information}
Naftali Tishby, Fernando~C Pereira, and William Bialek.
\newblock The information bottleneck method.
\newblock {\em arXiv preprint physics/0004057}, 2000.

\bibitem{tokmakov2017learning}
Pavel Tokmakov, Karteek Alahari, and Cordelia Schmid.
\newblock Learning motion patterns in videos.
\newblock In {\em Proceedings of the IEEE conference on computer vision and
  pattern recognition}, pages 3386--3394, 2017.

\bibitem{tokmakov2017learningb}
Pavel Tokmakov, Karteek Alahari, and Cordelia Schmid.
\newblock Learning video object segmentation with visual memory.
\newblock In {\em Proceedings of the IEEE International Conference on Computer
  Vision}, pages 4481--4490, 2017.

\bibitem{tran2015learning}
Du Tran, Lubomir Bourdev, Rob Fergus, Lorenzo Torresani, and Manohar Paluri.
\newblock Learning spatiotemporal features with 3d convolutional networks.
\newblock In {\em Proceedings of the IEEE International Conference on Computer
  Vision (ICCV)}, pages 4489--4497, 2015.

\bibitem{wang2017untrimmednets}
Limin Wang, Yuanjun Xiong, Dahua Lin, and Luc Van~Gool.
\newblock Untrimmednets for weakly supervised action recognition and detection.
\newblock In {\em Proceedings of the IEEE conference on Computer Vision and
  Pattern Recognition (CVPR)}, pages 4325--4334, 2017.

\bibitem{wang2016temporal}
Limin Wang, Yuanjun Xiong, Zhe Wang, Yu Qiao, Dahua Lin, Xiaoou Tang, and Luc
  Van~Gool.
\newblock Temporal segment networks: Towards good practices for deep action
  recognition.
\newblock In {\em European conference on computer vision}, pages 20--36.
  Springer, 2016.

\bibitem{wang2018videos}
Xiaolong Wang and Abhinav Gupta.
\newblock Videos as space-time region graphs.
\newblock In {\em Proceedings of the European conference on computer vision
  (ECCV)}, pages 399--417, 2018.

\bibitem{xu2017r}
Huijuan Xu, Abir Das, and Kate Saenko.
\newblock R-c3d: Region convolutional 3d network for temporal activity
  detection.
\newblock In {\em Proceedings of the IEEE international conference on computer
  vision}, pages 5783--5792, 2017.

\bibitem{xu2020g}
Mengmeng Xu, Chen Zhao, David~S Rojas, Ali Thabet, and Bernard Ghanem.
\newblock G-tad: Sub-graph localization for temporal action detection.
\newblock In {\em Proceedings of the IEEE/CVF Conference on Computer Vision and
  Pattern Recognition}, pages 10156--10165, 2020.

\bibitem{yang2019unsupervised}
Yanchao Yang, Antonio Loquercio, Davide Scaramuzza, and Stefano Soatto.
\newblock Unsupervised moving object detection via contextual information
  separation.
\newblock In {\em Proceedings of the IEEE Conference on Computer Vision and
  Pattern Recognition}, pages 879--888, 2019.

\bibitem{yuan2018marginalized}
Yuan Yuan, Yueming Lyu, Xi Shen, Ivor~W. Tsang, and Dit-Yan Yeung.
\newblock Marginalized average attentional network for weakly-supervised
  learning.
\newblock In {\em International Conference on Learning Representations (ICLR)},
  2019.

\bibitem{zeng2019graph}
Runhao Zeng, Wenbing Huang, Mingkui Tan, Yu Rong, Peilin Zhao, Junzhou Huang,
  and Chuang Gan.
\newblock Graph convolutional networks for temporal action localization.
\newblock In {\em Proceedings of the IEEE International Conference on Computer
  Vision}, pages 7094--7103, 2019.

\bibitem{zhao2017temporal}
Yue Zhao, Yuanjun Xiong, Limin Wang, Zhirong Wu, Xiaoou Tang, and Dahua Lin.
\newblock Temporal action detection with structured segment networks.
\newblock In {\em Proceedings of the IEEE International Conference on Computer
  Vision}, pages 2914--2923, 2017.

\bibitem{zhou2018temporal}
Bolei Zhou, Alex Andonian, Aude Oliva, and Antonio Torralba.
\newblock Temporal relational reasoning in videos.
\newblock In {\em Proceedings of the European Conference on Computer Vision
  (ECCV)}, pages 803--818, 2018.

\end{thebibliography}
}

\end{document}